\documentclass[11pt]{article}

\usepackage[preprint]{acl}

\usepackage{times}
\usepackage{latexsym}

\usepackage{amsmath}
\usepackage{amsthm}
\newtheorem{proposition}{Proposition}
\usepackage{subcaption}
\usepackage{amssymb}
\usepackage[T1]{fontenc}
\usepackage[utf8]{inputenc}

\usepackage{microtype}
\definecolor{darkgreen}{rgb}{0.0, 0.5, 0.0}
\usepackage{inconsolata}

\usepackage{graphicx}

\usepackage{booktabs}
\usepackage{multirow}
\usepackage{makecell}
\usepackage[table]{xcolor}

\title{TensorLens: End-to-End Transformer Analysis

via High-Order Attention Tensors}
\author{Ido Andrew Atad ~~~~~ Itamar Zimerman ~~~~~ Shahar Katz ~~~~~ Lior Wolf\\
Blavatnik School of Computer Science and AI, Tel Aviv University\\
\small{\texttt{\{idoatad,zimemran1,shaharkatz3\}@mail.tau.ac.il}},  \small{\texttt{ wolf@cs.tau.ac.il}}
}

\begin{document}
\maketitle
\begin{abstract}
Attention matrices are fundamental to transformer research, supporting a broad range of applications including interpretability, visualization, manipulation, and distillation. Yet, most existing analyses focus on individual attention heads or layers, failing to account for the model's global behavior. While prior efforts have extended attention formulations across multiple heads via averaging and matrix multiplications or incorporated components such as normalization and FFNs, a unified and complete representation that encapsulates all transformer blocks is still lacking. We address this gap by introducing TensorLens, a novel formulation that captures the entire transformer as a single, input-dependent linear operator expressed through a high-order attention-interaction tensor. This tensor jointly encodes attention, FFNs, activations, normalizations, and residual connections, offering a theoretically coherent and expressive linear representation of the model’s computation. TensorLens is theoretically grounded and our empirical validation shows that it yields richer representations than previous attention-aggregation methods. Our experiments demonstrate that the attention tensor can serve as a powerful foundation for developing tools aimed at interpretability and model understanding. Our code is attached as a supplementary.%
%Furthermore, we develop interpretability measures tools such as (i) attention-tensor distance metric allowing .... (ii)  attention-tensor entropy  ... built upon ... 
% Finally, we introduce an efficient computation scheme that facilitates practical adoption of our representation within the community.\idoa{we dont show the eff comp, maybe in app?}
\end{abstract}

\vspace{0.5em}
\hspace{.5em}
\includegraphics[width=1.25em,height=1.15em]{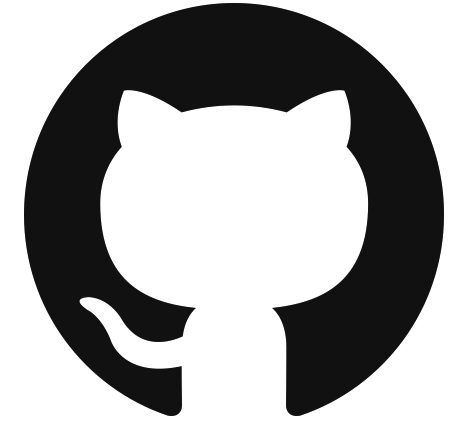}\hspace{.75em}
\parbox{\dimexpr\linewidth-7\fboxsep-7\fboxrule}{\url{https://github.com/idoatad/TensorLens}}
\vspace{-.5em}

% \vspace{-2pt}
\section{Introduction}
Transformer-based architectures~\citep{vaswani2017attention} have revolutionized deep learning by exhibiting remarkable scaling properties, enabling effective models with millions or even billions of parameters that can be trained on extensive datasets containing trillions of tokens. This advancement has led to breakthroughs that include large language models (LLMs) such as  ChatGPT~\citep{brown2020language}, Vision Transformers~\citep{dosovitskiy2020image}, Diffusion Transformers~\citep{peebles2023scalable}, and others. The core component of the Transformer responsible for capturing interactions between tokens is the self-attention mechanism.

{
\begin{figure}[t]
    \centering
    \includegraphics[width=1.0\linewidth]{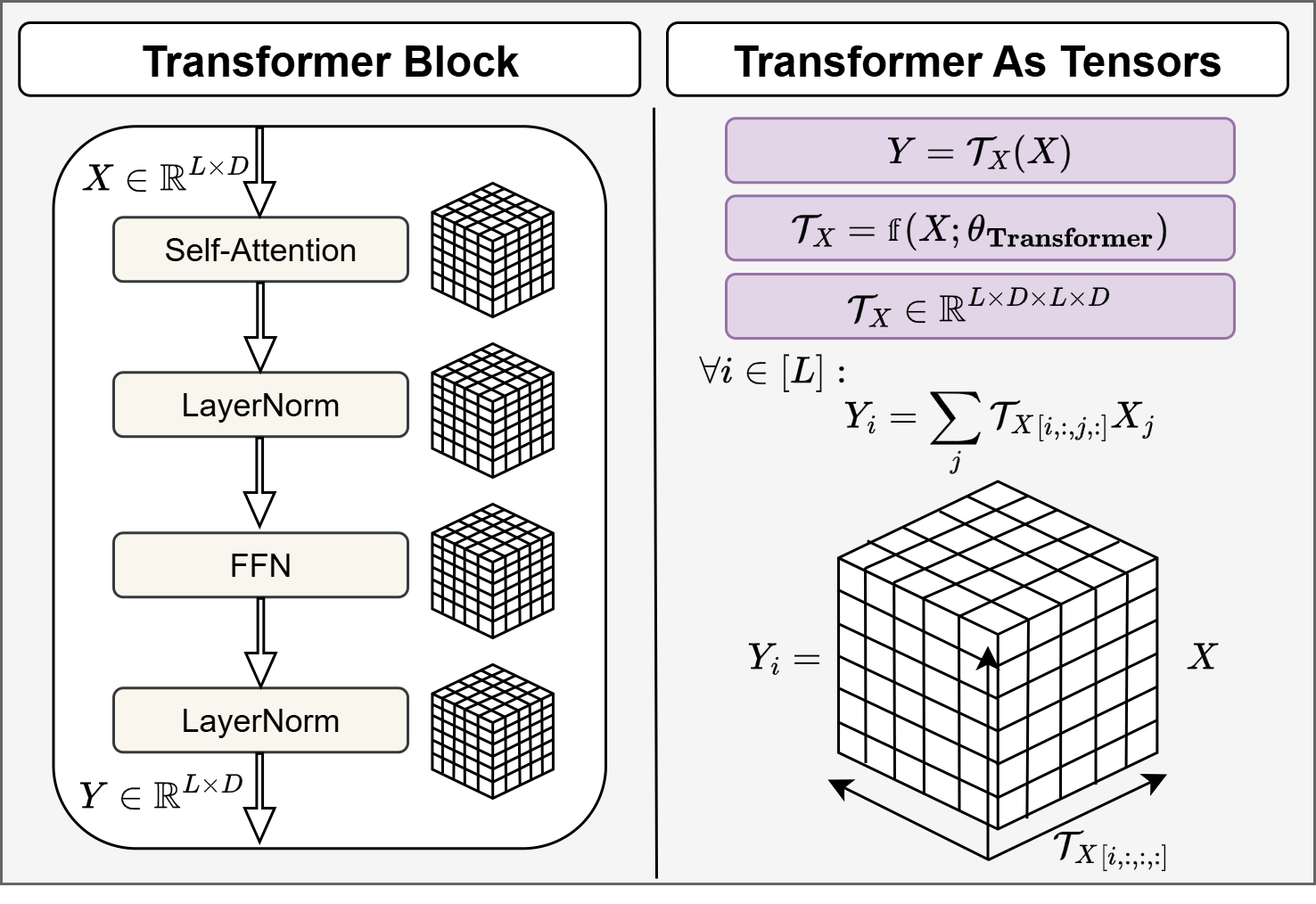}
    \caption{Transformers are re-formulated as data-controlled linear operators, characterized by an input-dependent high-order attention tensor $\mathcal{T}$. This formulation enables a unified self-attention representation that captures the entire Transformer architecture, including sub-components such as FFN layers, normalization, embedding layers, and residual connections.}
    \label{figure:method}
    \vspace{-8pt}
\end{figure}
}

Self-attention can be viewed as a data-controlled linear operator~\citep{poli2023hyena,massaroli2020dissecting} that is represented by an input-dependent attention matrix. Due to the row-wise softmax normalization, these attention matrices are somewhat interpretable, offering insight into how each layer updates the output representations through a weighted linear combination of input value vectors. As a result, attention matrices have been employed in a wide range of research domains, including (i) explainability and interpretability through attribution methods and model analysis~\citep{abnar2020quantifying, katz2024backward}, (ii) model editing and intervention techniques~\citep{chefer2023attend, katz-wolf-2025-reversed, alidetecting}, (iii) distillation and training techniques~\citep{touvron2021training,zhang2024lolcats}, (iv) inductive bias and regularization methods~\citep{li2018multi,attanasio2022entropy,zimerman2024viewing}, among others.

To push these applications further, substantial effort has been invested in developing extended representations of attention that go beyond individual attention matrices. A prominent example is the attention rollout technique~\citep{abnar2020quantifying}, which averages attention matrices across heads within the same layer and then integrates across the layers by applying multiplication. More recent approaches propose improved aggregation methods across heads. For example,~\citet{kobayashi2020attention} leverages the output projection layer to aggregate heads more precisely, and subsequent work incorporates the feed-forward layer into the formulation~\citep{kobayashi2023analyzing}. Additionally, in non-transformer models, implicit attention formulations have been introduced even for several architectures, including Mamba~\citep{ali2025hidden,dao2024transformers}, RWKV and Griffin~\citep{zimerman2025explaining}, and others%\lw{CITE ASLO THE MAMBA2 PAPER - Done}
. Following this line of work, we ask: What is the most comprehensive formulation that attention can encompass in Transformers? {Is it possible to represent the entire Transformer as a data-controlled linear operator that captures all of its parameters and is theoretically grounded, rather than relying on heuristically aggregated attention matrices?}

We fundamentally address this question by reformulating the entire Transformer model, including all of its components (feed-forward networks (FFNs), activation functions, LayerNorm, skip connections, embedding layers, and others) as a single data-controlled linear operator as visualized in Figure~\ref{figure:method}. A key insight of our work is that such a formulation requires high-order tensor %\lw{I THINK IT IS CALLED THE ORDER NOT THE DIM. True.}
attention tensors, not just matrices, to fully encompass the model’s behavior. Our formulation is theoretically grounded, and our empirical analysis shows that it better reflects the model than previously proposed attention forms.
Moreover, we demonstrate that the tensor structure can approximate linear relations~\cite{hernandez2024linearity} better than the matrix alternatives, underscoring its capacity to reveal LLM functionalities previously explored in mechanistic interpretability research.

% \textbf{Our main contributions are as follows:} (i) We propose a novel formulation based on high-order tensors that reformulates the entire Transformer model as a single data-controlled linear operator. (ii) We theoretically demonstrate that this formulation is more principled and precise than prior attention formulations, and it encompasses all model parameters. (iii) We empirically validate that our formulation better reflects the model's behavior through a series of perturbation tests. Finally, {(iv) We demonstrate that our tensor‑based attention formulation provides a robust foundation for mechanistic interoperability tools, such as approximating of linear relations from LLMs' embedding.}
%ido: replaced to make it slightly clearer, and added..
\textbf{Our main contributions are as follows:} (i) We introduce TensorLens, a novel high-order tensor formulation that represents the entire Transformer as a data-controlled linear operator, yielding generalized attention maps that can replace standard attention matrices and their cross-layer aggregations.
(ii) We provide theoretical justification showing that this formulation is principled, more precise than prior attention variants, and encompasses all model parameters.
(iii) We empirically show that TensorLens better reflects model behavior through perturbation-based evaluations.
Finally, (iv) we demonstrate that TensorLens provides a robust foundation for mechanistic interpretability tools, such as approximating linear relations from LLM embeddings.

\section{Background \& Related Work}
This section provides the scientific context for discussing our approach to precisely aggregating attention matrices via high-order tensors.

\subsection{Extended Attention Matrices\label{subsec:extensionAttentionMatrices}}
Due to their importance, extended formulations of attention matrices have been widely explored over the years. In particular,~\citet{kobayashi2020attention} demonstrated that attention analysis can be refined by incorporating the output projection matrix when analyzing Transformer heads. Additionally,~\citet{kobayashi2021incorporating} proposed a further refinement by incorporating the residual connections and normalization layers into the attention formulation, resulting in more precise formulation. These approaches were further extended by~\citet{kobayashi2023analyzing} who also incorporated the FFN sub-layer into the attention analysis. Moreover,~\citet{abnar2020quantifying} introduced the attention rollout technique, which aggregates attention weights across multiple layers by multiplying the per-layer attention matrices. The rollout method was applied by~\citet{modarressi-etal-2022-globenc} to aggregate the extended attention matrices of~\citet{kobayashi2021incorporating} across layers. Finally, most similarly to our work,~\citet{elhage2021mathematical} analyze 2 layer attention-only transformers using 4th order tensors to describe the end-to-end function of the model. Our approach builds on these works by proposing a more precise formulation that explicitly captures all Transformer blocks and their sub-components. 

\subsection{Attention as High-Order Tensors}
Several prior works have proposed architectures that extend the matrix-based self-attention mechanism to higher-order tensors~\citep{omranpour2025higher,ma2019tensorized,gao2020kronecker,zhang2025tensor}, primarily to enhance expressivity~\cite{sanford2023representational}. However, these approaches often come at the cost of reduced efficiency, prompting efforts to improve their computational performance~\citep{liang2024tensor}. While related, this line of work focuses on architectural modifications to the Transformer, rather than reinterpreting the vanilla self-attention mechanism through a tensor-based formulation, as done in this work.

\subsection{Explainability Attribution Methods}
Attribution methods aim to explain the decisions of neural networks (NNs) by quantifying the contribution of each neuron or input feature to the model’s output~\citep{das2020opportunities}. These tools are primarily used for interpretability and are crucial for making NNs more trustworthy and understandable~\cite{doshi2017towards}. Attribution can be either class-specific, where the explanation targets a particular output class (for example, why the model predicted ``cat'' over ``dog''), or class-agnostic, where the method provides a general explanation of the model's behavior regardless of any specific output~\citep{hassija2024interpreting}. While class-specific methods are valuable for understanding individual decisions, class-agnostic methods offer insights into the model’s global processing, emergent patterns, and internal representations. Both perspectives are complementary and play a central role in building explainable AI systems. 

Popular class-specific attribution methods include gradient-based techniques such as Input $\times$ Gradient~\citep{shrikumar2017learning,baehrens2010explain}, and Layer-wise Relevance Propagation (LRP)~\cite{bach2015pixel,achtibat2024attnlrp,bakish2025revisiting}. In contrast, common class-agnostic methods include activation maximization~\citep{erhan2009visualizing}, probing techniques~\citep{alain2016understanding}, and the extraction of attention maps~\cite{abnar2020quantifying}. This paper focuses on developing a class-agnostic explainability method for Transformers, based on a more generalized and insightful formulation of attention matrices via Tensors.

%%%%%%%%%%%%%%%%%%%%%%%%%%%%%%%%%%%%%%%%%%%%%%%%%%%%%%%%%%%%%
\section{Method: TensorLens} \label{sec:methods}
A standard Transformer architecture with $N$ layers and hidden representations $X^n\in \mathbb{R}^{L\times D}$ for any $n \in [N] $ is defined as follows: %
\begin{equation}\label{eq:BlockTransformer}
\forall n \in [N] : X^{n+1} = \text{Transformer}^{n}(X^{n})\,, 
\end{equation}
where each Transformer block is defined by:
\begin{equation}\label{eq:attnTransformer}
Z^{n} = \text{LayerNorm}^{n}(\text{Attention}^{n}(X^{n}) + X^{n}) \,,
\end{equation}
\begin{equation}\label{eq:FFNTransformer}
X^{n+1} = \text{LayerNorm}^{n}(\text{FFN}^{n}(Z^{n}) + Z^{n}) \,.
\end{equation}

\begin{figure*}
    \centering
    \includegraphics[width=1.0\linewidth]{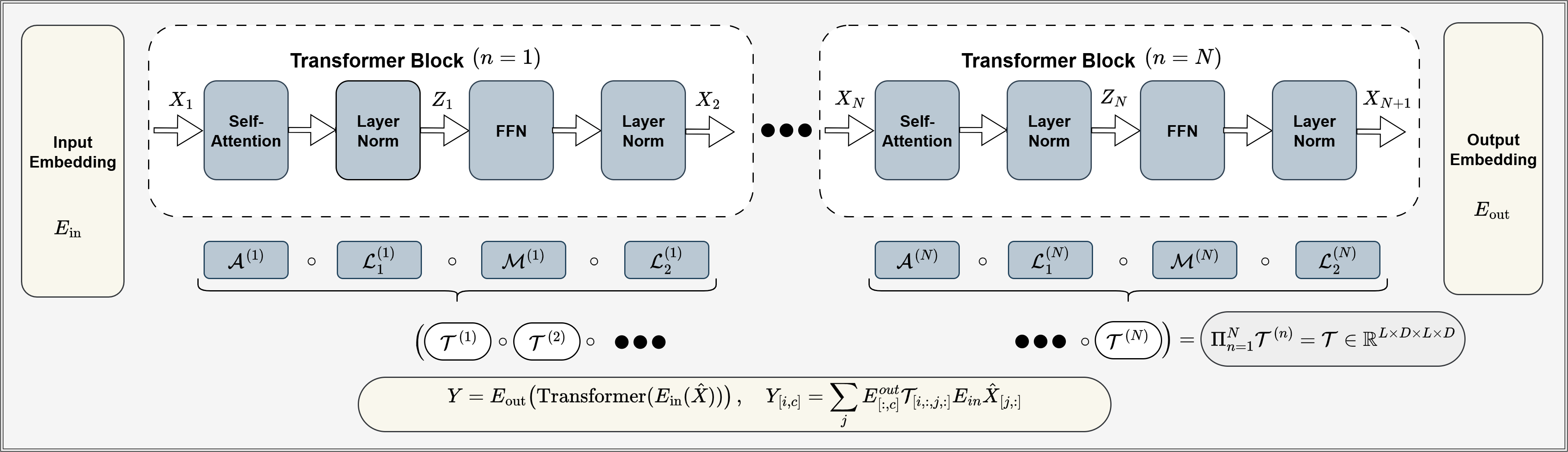}
    \caption{\textbf{Method:} A schematic visualization of our method, where each sub-component of the transformer architecture, including self-attention, LayerNorm, FFNs, input and output embedding layers, and the residual connection (which is omitted here for simplicity), is formulated as a data-control linear operator represented by high-order tensor in $\mathbb{R}^{L \times D \times L \times D}$. These tensors are composed into per-block tensors $\mathcal{T}^{(n)}$ for each layer $n \in [N]$, which are then used to construct the final linear operator % $\mathcal{T}$ 
    representing the entire Transformer.% \iz{Add embedding matrices.}
    } 
    \label{fig:levelOfTensor}
\end{figure*}
\noindent
Here, \text{LayerNorm} denotes the layer normalization operation \cite{ba2016layer}, \text{FFN} is the feed-forward layer, and \text{Attention} is the self-attention layer. The superscript $n$ indicates the signals, parameters or operations corresponding to the $n$-th layer, and each intermediate representation is a matrix in $\mathbb{R}^{L \times D}$, where $L$ is the sequence length and $D$ is the hidden dimension. We also assume that the input and output are multiplied by an embedding matrices $E_\text{in}$ and $E_\text{out}$.

% \vspace{-3pt}
\paragraph{Intuition.} Our key insight is that each sub-component of the Transformer can be represented as a data-controlled linear operator defined by a data-dependent matrix. However, while some components, such as attention, mix interactions between tokens, others, like the FFN, mix across dimensions. As a result, their combination cannot be represented by a single matrix. Instead, it requires a tensor-based operator to capture both types of interactions.

To materialize our insight, we show that each sub-layer in Transformers can be represented as a tensor-based data-control linear operator in Section~\ref{subsec:TensorBlock}, followed by how these tensors can be aggregated to represent each block and the entire model in Sections~\ref{subsec:TensorSingleBlock},~\ref{subsec:TensorModel}, and~\ref{subsec:Collapsing}. A schematic visualization of the method is presented in Figure~\ref{fig:levelOfTensor}.

\subsection{Prerequisites\label{subsec:Prerequisites}}
Our formulation of the Transformer as tensors builds on the following rules for vectorizing matrix operations and tensor calculations~\citep{itskov2007tensor}:

%begin{itemize}
    %\item
% \vspace{-3pt}
\paragraph{Bilinear Map.} A bilinear map $AXB$ can be vectorized using the Kronecker product $\otimes$ as:
    \begin{equation}\nonumber
\text{vec}\bigl[\underbrace{A}_{L\times L}\underbrace{X}_{L\times D}\underbrace{B}_{D\times D}\bigr]=\underbrace{\left(B^{\top}\otimes A\right)}_{LD\times LD}\underbrace{\text{vec}\left[X\right]}_{LD}\,.
\end{equation}
%    \item 

\vspace{-3pt}
\paragraph{Matrix Multiplication.} A matrix multiplication $XM$ can be vectorized as:
\begin{equation}\nonumber
\text{vec}\bigl[\underbrace{I_{L}}_{L\times L}\underbrace{X}_{L\times D}\underbrace{M}_{D\times D}\bigr]=\underbrace{\left(M^{\top}\otimes I_{L}\right)}_{LD\times LD}\underbrace{\text{vec}\left[X\right]}_{LD}\,.
\end{equation}

\vspace{-3pt}
\paragraph{Element-wise Hadamard Product.} An element-wise Hadamard product is vectorized as:
    \begin{equation}\nonumber
\text{vec}\bigl[\underbrace{H}_{L\times D}\odot\underbrace{X}_{L\times D}\bigr]=\underbrace{\text{diag}\left(\text{vec}\left[H\right]\right)}_{LD\times LD}\underbrace{\text{vec}\left[X\right]}_{LD}\,.
\end{equation}

\vspace{-3pt}
\paragraph{Tensor Contractions.}
For an input matrix $X\in\mathbb{R}^{L\times D}$ and a 4th order tensor $\mathcal{T}\in\mathbb{R}^{L\times D \times L \times D }$, we define the tensor contraction $\mathcal{T}\left(X\right) $ %\in\mathbb{R}^{L\times D}$ 
as:
    \begin{equation}
\forall i\in\left[L\right]:\,\,\mathcal{T}\left(X\right)_{\left[i,:\right]}=\sum_{j=1}^{L}\underset{D\times D}{\underbrace{\mathcal{T}_{\left[i,:,j,:\right]}}}\underset{D}{\underbrace{X_{\left[j,:\right]}}}\in\mathbb{R}^{D}\,.
    \end{equation}
Unfolding the tensor into a matrix $\mathcal{T}_{\text{mat}}\in\mathbb{R}^{LD\times LD}$, the vectorized tensor contraction follows
    \begin{equation}\label{eq:preq_mat_tensor} \text{vec}\left[\mathcal{T}\left(X\right)\right]=\mathcal{T}_{\text{mat}}\text{vec}\left[X\right]\,.
    \end{equation}
In the following sections we overload notation, referring to $\mathcal{T}_{\text{mat}}$ as $\mathcal{T}$.

\subsection{Block-by-Block Tensorization \label{subsec:TensorBlock}}
We now show how each sub-layer in the Transformer architecture (LayerNorm, self-attention, FFN, residual) can be ``Tensorized'' into a linear tensor form. For simplicity, in this section we omit superscripts and weight biases, derivation including biases is in Appendix \ref{sec:app_biases}.

\vspace{-3pt}
\paragraph{Tensorized Self-Attention.}
Recall that given an input $X$, the multi-head self-attention layer with $H$ heads is parameterized by key $W_{k,h}$, query $W_{q,h}$, value $W_{v,h}$, and output $W_{o,h}$ projections for each head $h \in [H]$, and is defined by:
\begin{equation}
\mathrm{Attn}(X)
    =\sum_{h=1}^{H} A_{h}\,X\,W_{v,h}\,W_{o,h}\,,
\end{equation}
\begin{equation}
A_{h} = \text{softmax}(Q_{h}K_{h}^{\!\top})\,,
\end{equation}
\begin{equation}
Q_{h} = X W_{q,h},\quad K_{h} = XW_{k,h}\,.
\end{equation}
Vectorising and grouping heads gives the following attention tensor $\mathcal{A}$:
\begin{equation}\label{eq:AttentionAsTensor}
\mathrm{vec}[\mathrm{Attn}(X)]=\underbrace{\sum_{h=1}^{H}\!\left(\left(W_{v,h}W_{o,h}\right)^{\top}\otimes A_{h}\right)}_{{\displaystyle \mathcal{A}}}\mathrm{vec}[X] \,.
\end{equation}

% \vspace{-3pt}
\paragraph{Tensorized LayerNorm.}
Recall that LayerNorm applies an affine transformation based on the input statistics and operates independently on each token:
\begin{equation}
\mathrm{LN}(X)=\gamma\!\odot\!\tfrac{X-\mu}
{\sigma}+\beta \,,
\end{equation}
where $\gamma$$\in\mathbb{R}^D$ and $\beta$ are learnable parameters, and $\mu$ and $\sigma \in\mathbb{R}^L$ are the per-token statistics , all broadcasted to match $X\in\mathbb{R}^{L\times D}$.  With pre-computed variance $\sigma^{2}$, the LayerNorm can be tensorized by:

\begin{equation}\label{eq:LayerNormAsTensor}
\mathrm{vec}\!\bigl[\mathrm{LN}(X)\bigr] = 
\mathrm{vec}\bigl[
    \mathrm{ diag }\!\bigl(\tfrac{1}{\sigma}\bigr)X
(I_{D}-\tfrac{\mathbf{1}\mathbf{1}^{\!\top}}{D})
      \mathrm{ diag} (\gamma)
\bigr] 
\end{equation}
\begin{equation}\nonumber
=\underbrace{\bigl[(I_{D}-\tfrac{\mathbf{1}\mathbf{1}^{\!\top}}{D})
      \mathrm{ diag} (\gamma)\bigr]^{\!\top}
      \!\otimes\!\mathrm{ diag }\!\bigl(\tfrac{1}{\sigma}\bigr)}_{\displaystyle\mathcal{L}}
      \mathrm{vec}[X]\,,
\end{equation}

where $(I_{D}-\tfrac{\mathbf{1}\mathbf{1}^{\!\top}}{D})\in\mathbb{R}^{D\times D}$ is the mean centering function in matrix form, with $\mathbf{1}\in\mathbb{R}^D$ a column vector of all ones.  

% \vspace{-3pt}
\paragraph{Tensorized FFN.}
Given an activation function $\phi$, the FFN is defined by two linear layers as follows:
\begin{equation}\label{eq:ffn}
\mathrm{FFN}(X) = \phi(XM_{1})M_{2}.  
\end{equation}

The element-wise activation can be converted to an input-dependent hadamard product $\frac{\phi(Z)}{Z}\odot Z$, and tensorized as:
\begin{equation}\label{eq:FFN_activation}
\mathrm{vec}\!\bigl[\phi(Z)\bigr] =\underbrace{\mathrm{diag}\!\left(\mathrm{vec}\left[\frac{\phi(Z)}{Z}\right]\right)}_{{\displaystyle \Psi}}\mathrm{vec}[Z]\,.
\end{equation}

Resulting in the full vectorized form of the FFN as follows:
\begin{equation}\label{eq:FFNAsTensor}
\mathrm{vec}\!\bigl[\mathrm{FFN}(X)\bigr]=\underbrace{\bigl(M_{2}^{\top}\otimes I_{L}\bigr)\Psi\bigl(M_{1}^{\top}\otimes I_{L}\bigr)}_{{\displaystyle \mathcal{M}}}\mathrm{vec}[X]\,,
\end{equation}
which is characterized by a tensor $\mathcal{M}$.

% \vspace{-3pt}
\paragraph{Tensorized Residual.}
For some sub-layer $g$, the residual connection can be written as:
\begin{equation}
    \mathbf{Y}_{\text{res}}=X+g(X)\,.
\end{equation}
Vectorizing this equation yields:
\begin{equation}\label{eq:residualAsTensor}
\mathrm{vec}\!\bigl[\mathbf{Y}_{\text{res}}\bigr]
=\bigl(I+\mathcal{G}\bigr)\mathrm{vec}\bigl[X\bigr]
       \,,
\end{equation}
where $\mathcal{G}$ is the tensor associated with $g$ (e.g.\ $\mathcal{A}$ or $\mathcal{M}$ for attention and FFN accordingly) and $\mathcal{I}\in\mathbb{R}^{LD\times LD}$ an identity matrix.

\subsection{Transformer Block as Tensor\label{subsec:TensorSingleBlock}}
% \lw{CHANGE THE ORDER SUCH THAT YOU DO NOT USE FURWARD DEFINITIONS } \iz{ I defined this equation at the beginning because of the 'Intuition' paragraph. I believe it makes the entire section easier to read with a clear understanding of the overall roadmap.} %
A Transformer block is defined in Eq.\ref{eq:BlockTransformer} and is obtained by stacking the sub-layers according to Eq.\ref{eq:attnTransformer} and Eq.~\ref{eq:FFNTransformer}. Thus, stacking the tensors obtained from the self-attention, residual, normalizations and FFNs as in
Eqs.~\eqref{eq:AttentionAsTensor},~\eqref{eq:LayerNormAsTensor},~\eqref{eq:FFNAsTensor}~\eqref{eq:residualAsTensor}, produces the tensor $\mathcal{T}^{n}$ associated with the $n$-th block: %
\begin{equation}\label{eq:BlockAtTensor}
\mathcal{T}^{n}=\mathcal{L}_{2}^{n}\left(\mathcal{M}^{n}+\mathcal{I}\right)\mathcal{L}_{1}^{n}\left(\mathcal{A}^{n}+\mathcal{I}\right)\,,
\end{equation}
for a post-layernorm block. For a similar derivation for a pre-layernorm block, see Appendix \ref{sec:app_biases}.

\subsection{Entire Transformer as Tensor\label{subsec:TensorModel}}
Given the tensor formulation of a single Transformer block in Section~\ref{subsec:TensorSingleBlock} (Eq.~\ref{eq:BlockAtTensor}), we now construct the full % Transformer
model as a composition of such block tensors. Let $\mathcal{T}^{n}$ denote the tensor representation of the $n$-th block, the entire Transformer function $\mathcal{F}$ can be expressed as a nested application of block tensors over the input sequence denoted as $X^0 =  X$, yielding the following recursive structure:
\begin{equation}
\mathcal{F}(X) = \mathcal{T}^{(N)} \circ \mathcal{T}^{(n-1)} \circ \cdots \circ \mathcal{T}^{(1)}(X)\,.
\end{equation}

Thus, the entire model is fully represented as a chain of high-order tensor transformations:
\begin{equation}\label{eq:transformer_as_tensor}
\mathrm{vec}[\mathcal{F}(X)] = \mathrm{vec}[\mathcal{T}(X)] = \left( {\prod_{n=1}^N \mathcal{T}^{n}} \right) \mathrm{vec}\bigl[X\bigr] \,,
\end{equation}
which completes the transition from individual layer operations to a unified tensor-based view of the full Transformer expressed by $\mathcal{T}(X)$.

\paragraph{Interpretation as Generalized Attention.}
We denote by $\mathcal{T}$ the linearized representation of the entire Transformer, expressed as a 4th-order tensor of dimensions $L \times D \times L \times D $, where $L$ is the sequence length and $D$ is the hidden dimension. This tensor captures the influence of each input token–channel pair on every output token–channel pair. Conceptually, $\mathcal{T}$ can be interpreted as a generalization of the conventional attention matrix to a \textit{higher-order attention tensor}, modeling both inter-token dependencies and intra-token (cross-channel) interactions. In the unvectorized form, each position $i\in \left[  L \right]$ in the output is obtained by a sum of linear transformations of the input, which are defined by slices of the overall tensor:
\begin{equation}\label{eq:full_nonvec}
\forall i \in [L]: \mathcal{F}\left(X\right)_{\left[ i,:\right]}=\sum_{j}\underset{D\times D}{\underbrace{\mathcal{T}_{\left[i,:,j,:\right]}^{}}}\underset{D}{\underbrace{X_{\left[ j,:\right]}^{\top}}}\,.
\end{equation}

Our goal with generalized attention is not to propose a new architecture, but to formulate attention in a way that yields a representation encapsulating more components and computations, following the line of work described in Section~\ref{subsec:extensionAttentionMatrices}. Importantly, our method is not limited to end-to-end linearization alone. By restricting the composition to a chosen subset of layers or heads, we can obtain generalized attention matrices at any desired granularity.
%{\color{blue} TBD. Interpretation as specialized Jacobian
% {\color{blue} Ido: could you rewrite this: the motivation here is the relation to the full jacobian.In our method we get a tensor that is a linear operation that reconstructs the output exactly, while the jacobian  does not, but the jacobian is the best linear approximation in the sense that the bound here will be $\Vert \epsilon \Vert_2$  for a small enough $\epsilon$. So the motivation here is to show that we can bound the "price" we pay in our method for getting the exact reconstruction (actually the bound is very loose but emperically small}

% Instead, i think we should talk here about how T can be understoof as a specialized jacobian, that by detaching the gradient on ffn activations, attention matrices, an ln std, can achieve the fact that the output is exacly reconstructed (instead of approx in the jacobian)
%}
%

% \lw{BE SPECIFIC ON ORDER AND DIMENSION SHERE. Solved}

\subsection{From Tensor to Matrix by Collapsing\label{subsec:Collapsing}}
%\paragraph{From Tensor to Matrix by Collapsing} 
While the tensor representation provides a richer and more comprehensive view of Transformer computations, it is often less interpretable and more difficult to visualize due to its 4th-order structure and the sheer number of elements ($L^2 D^2$ in total). To address this, we propose a simple yet effective technique for collapsing the tensor into a more compact, matrix-like form, akin to the standard attention matrix. Specifically, we reduce the $D \times D$ channel dimensions using the following three approaches: \textbf{(i) Norm over feature dimensions:} By taking the norm of the dimension related to the channel as follows:
    \begin{equation}\label{eq:tensorAggrNorm}
    \forall i,j\in[L]:T_{i\leftarrow j}^{\text{Norm}}=\left\Vert \mathcal{T}_{\left[i,:,j,:\right]}\right\Vert _{2} ,
    \end{equation}
resulting in a matrix $T^{\text{Norm}} \in \mathbb{R}^{L \times L}$. \textbf{(ii) Projection using output and input embedding vectors:} Let $X^0, X^N \in \mathbb{R}^{L \times D}$ be the hidden states inserted and extracted from the Transformer layers. We contract over the channel dimensions using $X^0, X^N$ as both input and output projection weights:
% \begin{equation}
%     \forall i,j \in [L] :
% \end{equation}
% \begin{equation}\nonumber
%         \tilde{T}_{i,j} = \sum_{d_1=1}^{D} \sum_{d_2=1}^{D} X^{0}_{d_1} \, \mathcal{T}_{i,j,d_1,d_2} \,  X^N_{d_2} =  {X^0}^\top \mathcal{T}_{i,j}  X^N
% \end{equation}
%
\begin{equation}\label{eq:tensorAggrEmbedProj}
\forall i,j\in[L]:T_{i\leftarrow j}^{\text{IO}}=X_{\left[i,:\right]}^{N}\mathcal{T}_{\left[i,:,j,:\right]}X_{\left[j,:\right]}^{0} \,.
\end{equation}
Following Eq.~\ref{eq:full_nonvec}, taking an inner product with $X^N_{\left[ i,:\right]}$ on both sides yields: 
\begin{equation}\label{eq:sum_T_norm}   {X_{[i,:]}^{N}}^{\top}X_{\left[i,:\right]}^{N}=X_{\left[i,:\right]}^{N}\sum_{j}\mathcal{T}_{\left[i,:,j,:\right]}X_{\left[j,:\right]}^{0}\,,
\end{equation}
\begin{equation}\nonumber
\left\Vert X_{\left[i,:\right]}^{N}\right\Vert ^{2}=\sum_{j}T_{i,j}^{\text{IO}}\,,
\end{equation}
meaning $T_{i,j}^{\text{IO}}$ reflects the contribution of the input $X^0_{\left[ j,:\right]}$ to the output $X^N_{\left[ i,:\right]}$.

This approach can be applied to the entire Transformer or to a selected subset of blocks.
\textbf{(iii) Class-specific projection using output embedding matrices:} For a chosen output class/token $c$, let $E_{\left[:,c\right]}^{out}\in\mathbb{R}^D$ be the corresponding column in the classification head/unembedding matrix, we contract the tensor over the channel dimensions using $X^0$, $E_{\left[:,c\right]}^{out}$ to get:
\begin{equation}\label{eq:TransformerAndEmddingAsTensor}
\forall i,j\in\left[L\right]:T_{\left(c,i\right)\leftarrow j}^{\text{CLS}}=E_{\left[:,c\right]}^{out}\mathcal{T}_{\left[i,:,j,:\right]}X_{\left[j,:\right]}^{0}\,.
\end{equation}
Similarly to Eq.~\ref{eq:sum_T_norm}, for each class $c$ and output position $i$, the input contributions $T_{\left(c,i\right)\leftarrow j}^{\text{CLS}}$ sum up to the logit of the class $c$ (excluding biases, see Appendix \ref{sec:app_biases}).

% For the projection, one can use either the actual embeddings inserted into or extracted from the Transformer, or alternatively, the embedding matrices $E_{\text{in}}$ and $E_{\text{out}}$. This results in the following formulation:
% %
% \begin{equation}\label{eq:TransformerAndEmddingAsTensor}
% \forall i,j \in [L] : \quad \tilde{T}_{i,j} = E_{\text{in}}^\top \mathcal{T}_{i,j} E_{\text{out}} \,.
% \end{equation}
% {\color{blue} TODO: fix this also as there is a subset of the embedding and output matrix}
Eq.~\ref{eq:TransformerAndEmddingAsTensor} provides a comprehensive view of the Transformer as a linear operator. It encapsulates all model parameters, including the Transformer and embedding layers, and serves as a direct local approximation of the full Transformer computation. As the first formulation that explicitly captures all model parameters within a unified tensor-based representation, it offers a principled foundation for analyzing and interpreting Transformer computations through the lens of high-order linear operators.

\subsection{Theoretical Analysis}
Our method relies on a linearization of the \textit{entire} Transformer computation at a given input. A natural question arises: how well does this linearization approximate the original function locally? We address this in Proposition~\ref{proposition:senitivity}, where we provide a data and model-dependent bound on the forward approximation error. It is important to note that alternative methods, which either do not incorporate all model parameters in their formulation or avoid tensor-level operations, are generally not capable of producing such bounds. Even when they can be applied, the resulting approximations are typically significantly looser. %\idoa{can we say this? we dont show emperical results}\iz{I think it makes sense, so yes. but it's a bit vague. Lior?}\idoa{Ido: i can run a quick experiment to get typical spectral norms of the tensor if it helps}
\begin{proposition}\label{proposition:senitivity} 
The \textbf{approximation error} of the tensor $\mathcal{T}_X$ computed on input $X$, when evaluating the transformer function $\mathcal{F}$ at $\left(X+\epsilon\right)$ is bounded by:
\begin{equation}
{\left\Vert \mathcal{T}_{X}\left(X+\epsilon\right)-\mathcal{F}\left(X+\epsilon\right)\right\Vert _{2}}\,\leq
\end{equation}
\begin{equation}\nonumber
\left\Vert \mathcal{T}_{X}\right\Vert _{2}\left\Vert \epsilon\right\Vert _{2}+\left\Vert \mathcal{F}\left(X+\epsilon\right)-\mathcal{F}\left(X\right)\right\Vert _{2}\,,
\end{equation}
where $\left\Vert \mathcal{T}_{X}\right\Vert _{2}$ is bounded by constants of the transformer weights.
\end{proposition}
The complete proof is provided in Appendix  ~\ref{sec:app_approx_bound}. The core intuition is that each sub-component of the Transformer can be linearly approximated using tensor operations, enabling the error to be bounded by recursively applying standard first-order linear approximation techniques, which can be composed across layers to yield a global %approximation
bound.

% \subsection{Computational Complexity Analysis}
% \iz{Ido, do we have something interesting to write here? Not a must.}
%%%%%%%%%%%%%%%%%%%%%%%%%%%%%%%%%%%%%%%%%%%%%%%%%%%%%%%%%%%%%
\begin{figure*}[t]
    \centering
        \includegraphics[width=0.95\textwidth]{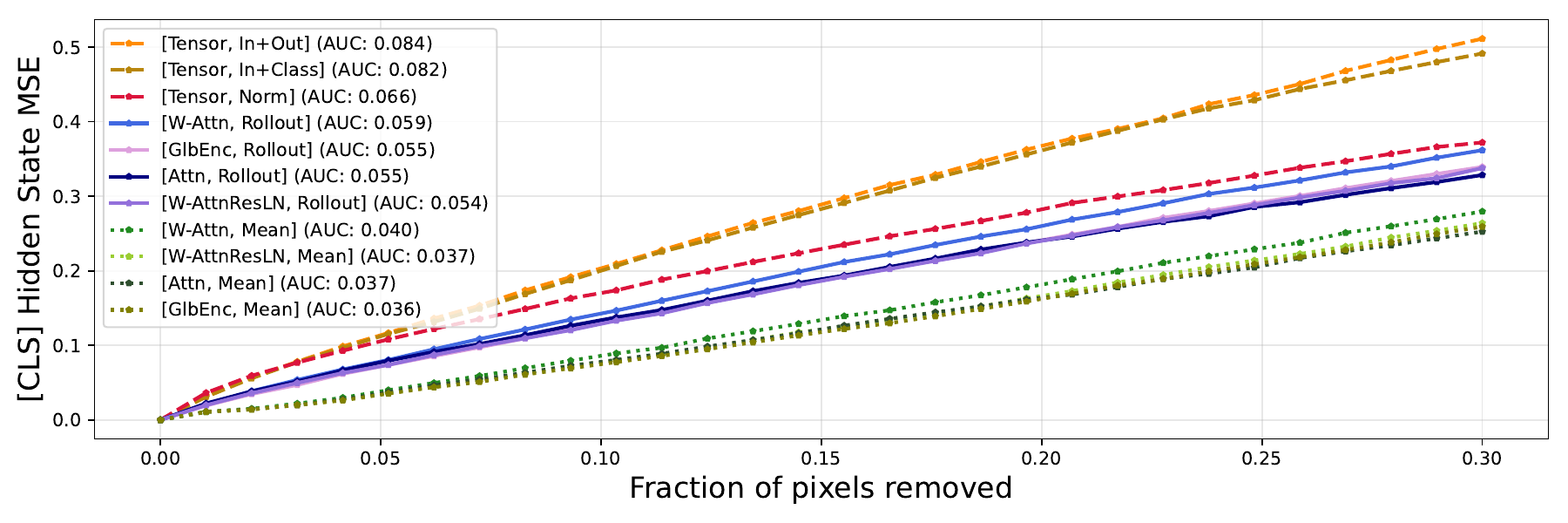}
    \caption{\textbf{Perturbation Tests in Vision:} Effect of perturbations on final hidden representations of DeiT-Base. Measured by the mean squared error between the last hidden-state of the [CLS] token in the original and perturbed input (higher is better).}
    \label{fig:pertubationVision_DEIT-BASE}
\end{figure*}

\begin{figure*}[h]
    \centering
        \includegraphics[width=0.95\textwidth]{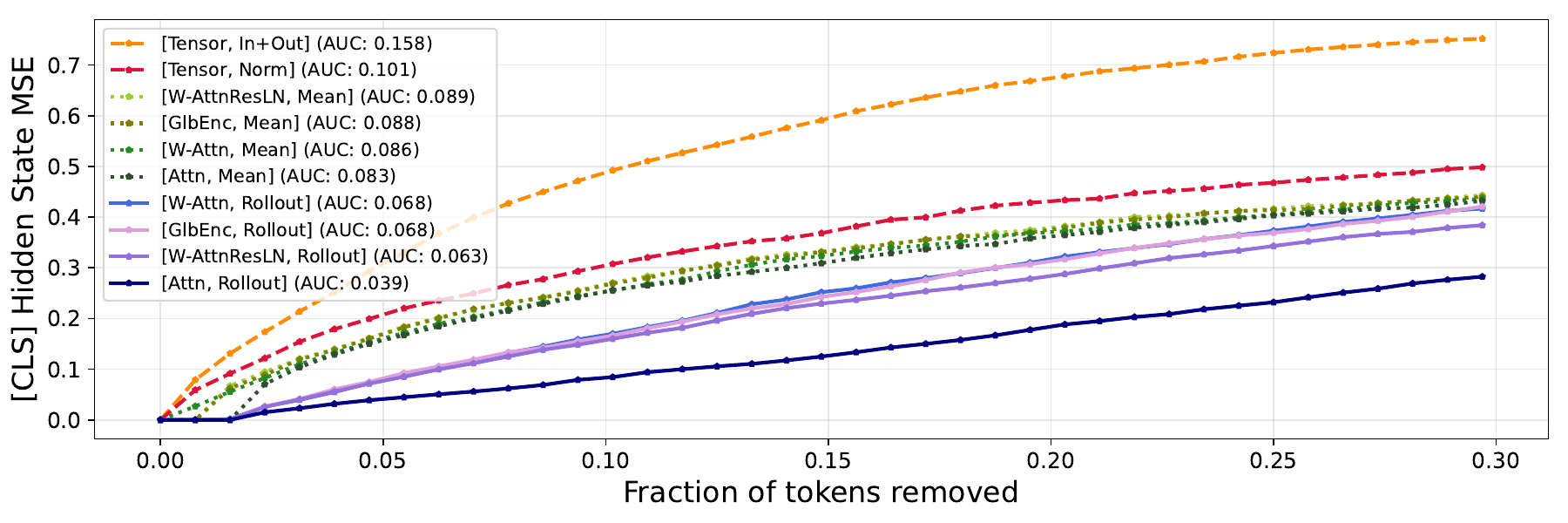}

\caption{\textbf{Perturbation Tests in NLP: } Effect of token perturbations on final hidden representations of BERT-Base.}
\label{fig:pertubationIMDB_BERT}
\end{figure*}

\section{Experiments}\label{sec:experiments}
We empirically assess the representation power of our tensor formulation as a proxy for Transformer behavior, comparing it to other attention aggregation techniques via perturbation tests in Section~\ref{subsec:pertubationTests}. {Then, in Section~\ref{subsec:arpproxLinear}, we demonstrate that the tensor representation is a valuable tool for mechanistic interpretability and model understanding.} %showing that it can surprisingly well approximate linear relationships. {\color{red}Ido or Shahar, Please rewrite the last sentence.}}

\subsection{Perturbation Tests\label{subsec:pertubationTests}}
To assess the representational power of our attention aggregation method, we adopted an input perturbation scheme similar to~\citet{chefer2021transformer,ali2022xai}. This evaluation strategy gradually masks input tokens in the order determined by their computed relevance scores. When the highest-scoring tokens are masked first (positive perturbation), we expect the model’s accuracy to rapidly decline. We assess explanation quality using the Area Under the Curve (AUC) metric, which captures the model's accuracy as a function of the percentage of masked input elements, ranging from 0\% to 30\%.

As baselines, we use eight %attention 
aggregation variants that combine two methods for cross-layer aggregation and four methods for intra-layer aggregation. For cross-layer aggregation, we apply either multiplicative composition, as in Attention Rollout~\cite{abnar2020quantifying} (``Rollout''), or simple averaging (``Mean''). For intra-layer aggregation, we use the following four methods: (i) averaging of attention matrices (``Attn''), (ii) value-weighted attention as proposed by~\citet{kobayashi2020attention} (``W. Attn''), (iii) value-weighted attention that also includes the residual connection and LayerNorm as proposed by~\citet{kobayashi2021incorporating} (``W. AttnResLN''), and (iv) a global encoding variant (``GlbEnc'') that further incorporates the second LayerNorm into the formulation~\citep{modarressi-etal-2022-globenc}. We compare these class-agnostic baselines with the variants defined in Eqs.~\ref{eq:tensorAggrNorm} and~\ref{eq:tensorAggrEmbedProj}, denoted as ``Tensor,Norm'' and ``Tensor,In+Out'', respectively.%\idoa{add that we compare only to class-free methods}\iz{added}.

% \vspace{-3pt}
\paragraph{Perturbation in Vision.}
In the vision domain, we evaluate our methods using %a ViT variant (
DeiT by~\citet{touvron2021training}, on the ImageNet-1K test set, considering both the base and small model sizes. Results for the base model are shown in Fig.~\ref{fig:pertubationVision_DEIT-BASE}. As can be seen, across all perturbation levels, tensor-based aggregation methods consistently outperform the baselines. When incorporating both input and output embeddings ('Tensor,In+Out'), the total AUC exceeds 0.82, and reaches 0.66 when using the tensor norm ('Tensor,Norm'). In contrast, all aggregation methods that are not based on tensors fall below 0.6, highlighting the superior robustness of our formulation. The results for DeIT-small, which follow a similar trend are provided in Appendix~\ref{sec:appendix_perturbation}. %\lw{YOU CANNOT REFER TO AN IMAGE IN THE APPENDIX ONLY TO AN APPENDIX SECTION The results for DeIT-small, which follow a similar trend are provided in Appendix~\ref{sec:appendix_perturbation}.} % in Figure~\ref{fig:pertubationVision_DEIT_SMALL} at the Appendix, we provide the results for DeiT-small, showing a similar trend.

\begin{figure*}[h!]
    \centering
    \begin{subfigure}[b]{0.32\textwidth}
    \centering
    \includegraphics[width=\linewidth]{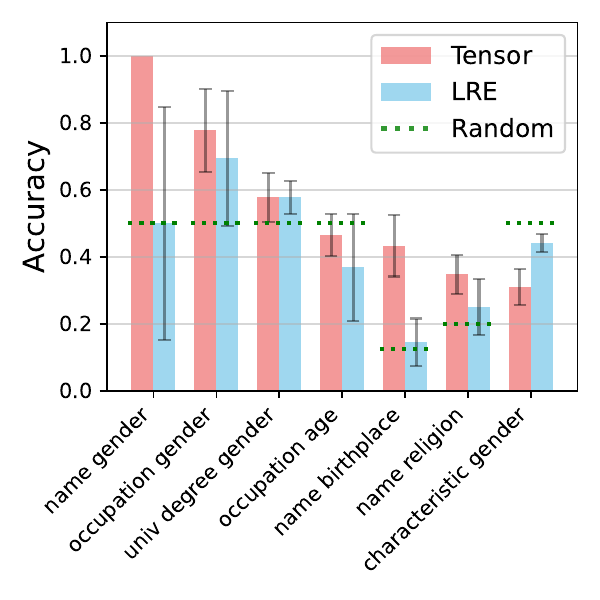}
    \caption{Bias}
    \end{subfigure}
    % \hfill
    \begin{subfigure}[b]{0.32\textwidth}
    \centering
    \includegraphics[width=\linewidth]{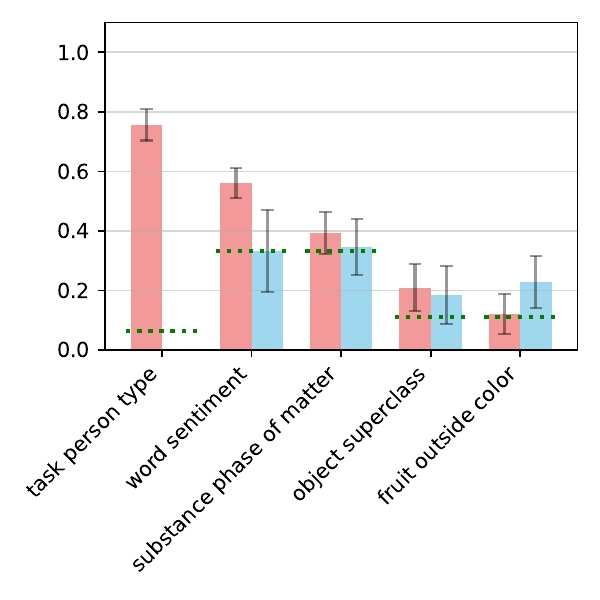}
    \caption{Common Sense}
    \end{subfigure}
    % \hfill
    \begin{subfigure}[b]{0.32\textwidth}
    \centering
    \includegraphics[width=\linewidth]{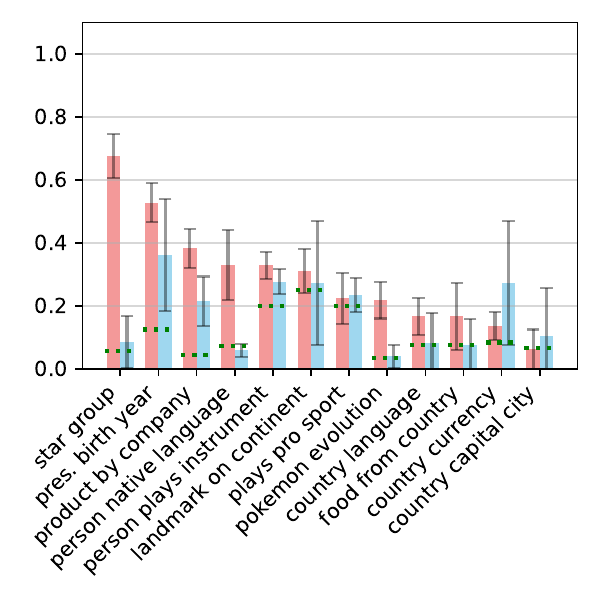}
    \caption{Factual}
    \end{subfigure}
     \vspace{-3pt}
    \caption{\textbf{Relation Decoding:} Accuracy relative to original model computation, for different relation categories on Pythia-1B, with $m=3$ training samples per relation. Results are averaged across 6 train-test splits, with standard deviation shown in error bars. Random baselines shown as horizontal dashed lines.}
    \label{fig:relation_pythia_1b}
\end{figure*}

\vspace{-3pt}
\paragraph{Perturbations in NLP.}
In the NLP domain, we evaluate our method across several models on sequences of length 128, including both encoder-only and decoder-only architectures. For the encoder-only setting, we conduct experiments with BERT~\citep{lu2019vilbert} and RoBERTa~\citep{liu2019roberta} on the IMDB dataset. Results for BERT are shown in Figure~\ref{fig:pertubationIMDB_BERT}, where tensor-based aggregation consistently outperforms all baselines across all perturbation levels. When incorporating both input and output embeddings ('Tensor,In+Out'), the total AUC exceeds 0.158, and reaches 0.101 when using only the tensor norm ('Tensor,Norm'). In contrast, all non-tensor aggregation methods fall below 0.09, underscoring the superior robustness of our formulation. Moreover, in Appendix~\ref{sec:appendix_perturbation}, we also report results for RoBERTa in Table~\ref{tab:Pythia-1B}, as well as for the more recent ModernBert \citep{modernbert} and Gemma3 \citep{team2025gemma} in Table~\ref{tab:bert_gemma}. In these figures, the observed pattern closely aligns with that of BERT.

When evaluating decoder-only models, the results are less clear-cut. In this setting, we tested several LLMs, including Pythia-1B~\citep{biderman2023pythia}, Pico-570M~\citep{martinez2024tending}, and Phi-1.5~\citep{textbooks2}, using the WikiText-103 dataset. As shown in Table~\ref{tab:Pythia-1B} (Appendix~\ref{sec:appendix_perturbation}), although our method consistently achieved the top or second-best AUC scores across benchmarks, the overall findings are less conclusive. One possible explanation is that auto-regressive language models are trained to predict the next token and therefore tend to exhibit inherently local behavior~\citep{fang2024wrong}. This characteristic may reduce the informativeness of perturbation-based evaluations, making the results appear less definitive. %\lw{IT LOOKS LIKE YOU HIDE THE RESULTS IN THE APPENDIX.% how ABOUT A SMALL TABLE OF AUC?
%}\iz{We did "hide" these results, as the table is quite large and not clear-cut —see the appendix.}

%%%%%%%%%%%%%%%%%

%%%%%%%%%%%%%%%%%%%

% \begin{figure*}[t]
%     \centering
%     \begin{subfigure}[b]{0.328\textwidth}
%     \centering
%     \includegraphics[width=\linewidth]{plots/relation/mean_per_layer/bias.pdf}
%     \caption{Bias}
%     \end{subfigure}
%     \hfill
%     \begin{subfigure}[b]{0.328\textwidth}
%     \centering
%     \includegraphics[width=\linewidth]{plots/relation/mean_per_layer/common_sense.pdf}
%     \caption{Common Sense}
%     \end{subfigure}
%     \hfill
%     \begin{subfigure}[b]{0.328\textwidth}
%     \centering
%     \includegraphics[width=\linewidth]{plots/relation/mean_per_layer/factual.pdf}
%     \caption{Factual}
%     \end{subfigure}
%     \caption{\textbf{Relation Decoding Using Intermediate Hidden States:} Mean accuracy per relation category, while computing Tensor and LRE \cite{hernandez2024linearity} approximations using varying initial layers (12 for Pythia-1B).}
%     \label{fig:relation_layers_mean}
% \end{figure*}

\subsection{Approximation of Relation 
Decoding\label{subsec:arpproxLinear}}
The tensor formulation $\mathcal{T}$, which we uncover from the forward pass of the model, mathematically describes a linear transformation between tokens in the same sentence. In this section we evaluate the quality of our method as a local approximation for the Transformer computation through the lens of \emph{linear relation decoding}. Introduced by \citet{hernandez2024linearity}, linear relation decoding examines sets of relations, such as \emph{``A \underline{teacher} typically works at a \underline{school}''}, composed of triplets $(s,r,o)$ connecting a subject $s$ to an object $o$ via relation $r$. 
\citet{hernandez2024linearity} illustrate how to produce transformation between tokens embeddings, such as ones that output \emph{``school''} for \emph{``teacher''} or \emph{``hospital''} for \emph{``doctor''}. Their method was based on approximating the Jacobian matrix of the model's prediction relative to the subject token ``s''.
Since our tensor formulation is a multi-linear transformation that describes such input-output relations, our goal is to examine to what extend it can match the linear representation's performances of \citet{hernandez2024linearity} which were tailored for this task.

In order to create a per-relation transformation, we compute the mean tensor extracted from $m$ examples $X_i=(s_i,r,o_i)$ of a relation $r$:
\begin{equation}\label{eq:mean_T_relation}  \widetilde{\mathcal{T}_r}=\frac{1}{m}\sum_{i=1}^{m}\mathcal{T}_{X_{i}}\,,
\end{equation}
and measure the similarity of the tensor function $\widetilde{\mathcal{T}_{r}}\left(X\right)$ to that of the original model, on a held-out test set of subject-object pairs of the same relation.  

% \paragraph{Experimental Setup} We compare our method to that of \citep{hernandez2024linearity}, which performs a similar mean value approximation of the model, using the jacobian matrix of the model's prediction relative to embedding of the subject token $s$.{\color{red} do i need to talk about their method at all?}
% Following the same experimental setup, while calculating the mean tensor for example $X_i$, 
For experimental setup, we follow \citet{hernandez2024linearity} and prepend each example $X_i$ in the mean calculation with the remaining $m-1$ train examples as few-shot examples, so that the model is more likely to generate the answer $o$ given a $s$ under the relation $r$ over other plausible tokens. Further experimental details are described in Appendix ~\ref{sec:appendix_reltion_relation_decoding}. We report the approximation accuracy as the percentage of examples in which the top-predicted object $o$ matches the original output.
%We report the accuracy of the approximation as the percentage of examples where the top object $o$ predicted %by our method 
% is the same as the original output.

 As seen in Figure~\ref{fig:relation_pythia_1b}, approximating the model's computation using our tensor method achieves higher accuracy than the LRE baseline of ~\citet{hernandez2024linearity} on most relations examined.
In some tasks, such as \textit{occupation-age}, we found both methods to achieve results close to that of a random guess, which we associate with the inherent limitation of describing the model's internal processes solely via linear transformations of the input.

Overall, it is evident that our multi-linear approximation provides better capacity than previous linear methods to describe the function of the \textit{entire} model as a whole. We find these results to strengthen our claim that the tensor formulation reflects the model's internal representations.

% Similarly to ~\citet{hernandez2024linearity}, our method can be applied using either the tensor of the entire model, or from a certain layer $n\in N$ onward. The mean tensor will be defined as \begin{equation}\label{eq:mean_T_relation_int_layer}  \widetilde{\mathcal{T}_{r}}^{n:N}=\frac{1}{m}\sum_{i}\left({\prod_{\mu=n}^{N}\mathcal{T}_{X_{i}}^{(\mu)}}\right)\,,
% \end{equation}
% and the model output is approximated as $\widetilde{\mathcal{T}_{r}}^{n:N}\left(X^{n}\right)$, using the intermediate hidden-states $X^n$ of the samples rather than the input embeddings.
% In Figure ~\ref{fig:relation_layers_mean} we compute both the tensor and LRE with varying initial layers. The accuracy of both methods increases throughout the layers, with our's typically being higher in the earlier layers, while out-performed in later layers. Additional per-relation results for this experiment are provided in Figure~\ref{fig:relation_layers_examples} of the appendix. 

% \begin{figure*}[t]
%     \centering \includegraphics[width=0.99\textwidth]{plots/relation/in_layer_0_all_relations_pythia1b.pdf}
%     \caption{\textbf{Relation Decoding Using Input Embeddings:} Accuracy relative to original model computation, for different relation categories on Pythia-1B, with $m=3$ training samples per relation. Results are average across 6 train-test splits, with standard deviation shown in error bars. Random baseline per relation shown as horizontal dashed lines.}
%     \label{fig:relation_pythia_1b}
% \end{figure*}

% \paragraph{Tensor Entropy??}

% \subsection{Ablations}

\section{Conclusions}
This work presents a technique for aggregating attention matrices across both Transformer blocks and all sub-components within each block. The resulting formulation is theoretically grounded and more comprehensive than prior approaches and it is based on representing the Transformer as a high-order data-controlled linear operator. %We anticipate that 
This formulation captures the internal interactions of the model, including contributions from components such as the FFN, embeddings, LayerNorm, and others. Practically, we emphasize that this formulation can be used as a drop-in replacement for attention matrices and their aggregations, in order to enhance many existing interpretability, analysis, and intervention techniques. An example of direct application to mechanistic interpretability and model understanding is demonstrated in our relation-based analysis. %These attention representation can allow researchers gain a deeper understanding of the dependencies learned by the model and support the development of improved task-agnostic explainability tools.

\section*{Limitations}
While TensorLens offers a more precise formulation of the Transformer through a self-attention-based representation compared to prior work, it has several limitations. First, some of the linearization techniques are chosen for their simplicity rather than being derived from the intrinsic properties of optimally approximated tensors. For example, this includes the activation decomposition in Eq.~\ref{eq:FFN_activation}. Second, the high-order tensor representation is GPU-memory intensive. We partially mitigate this with a memory-optimized computation method as described in Appendix \ref{sec:appendix_mem_eff_compute}, however, our experiments are limited to models up to 1B parameters and moderate input lengths. Third, although the formulation is comprehensive and practical for visualization and model interpretation, the full potential of the tensor-based approach remains underexplored. In particular, it opens the door to new perspectives on rank collapse, sparsity, and training dynamics through the lens of tensor properties.

\section*{Ethics Statement}
This work focuses on developing a theoretically grounded and interpretable representation of Transformer-based models via high-order attention tensors. Our research does not involve human subjects, personally identifiable data, or the generation of potentially harmful content. All evaluations are conducted on publicly available datasets such as ImageNet, IMDB, and WikiText-103, adhering to their respective licenses and intended usage.

We acknowledge that improved model interpretability tools, such as those proposed in this work, may be used both to enhance trust in machine learning systems and to expose or exploit model vulnerabilities. We believe that the positive implications --- such as greater transparency, accountability, and error analysis --- outweigh potential misuse. Nonetheless, we encourage responsible use of our methods in alignment with ethical AI principles.
% \section*{Acknowledgments}

% Bibliography entries for the entire Anthology, followed by custom entries
%\bibliography{anthology,custom}
% Custom bibliography entries only
\bibliography{custom}

\appendix
\newpage
\onecolumn
\section{Additional Perturbation Experiments\label{sec:appendix_perturbation}}
\begin{figure*}[h]
    \centering
            \includegraphics[width=0.9\textwidth]{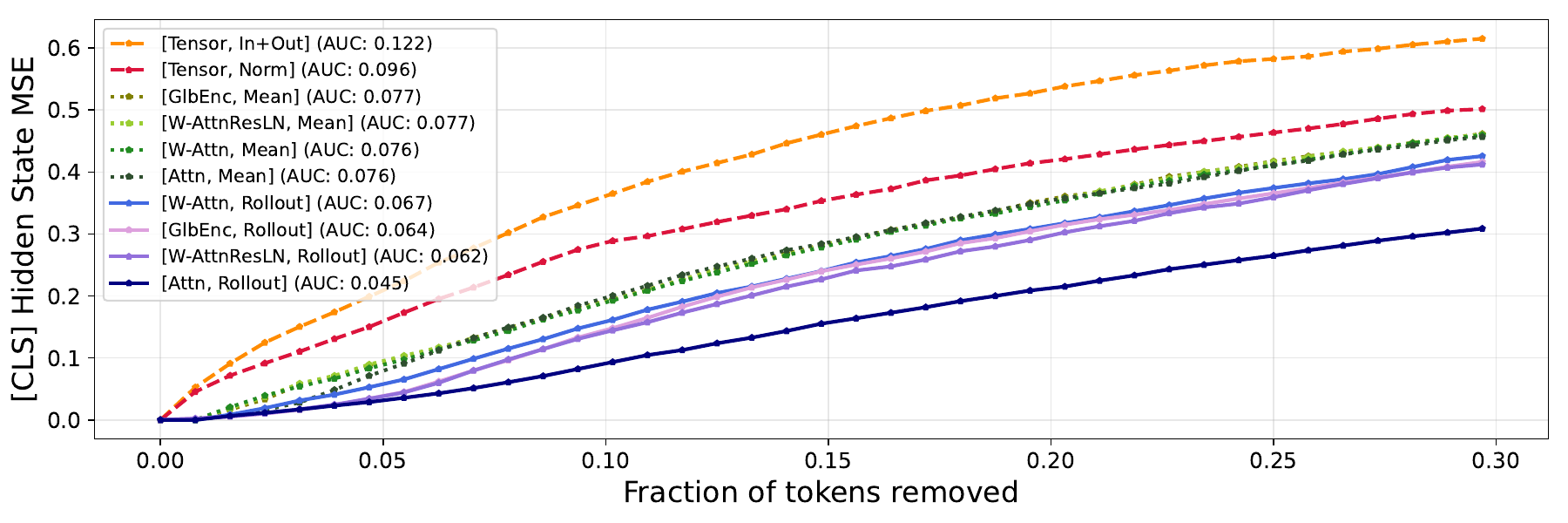}
\caption{\textbf{Perturbation Tests in NLP: } Effect of token perturbations on final hidden representations of RoBERTa-Base. Measured by the mean squared
error between the last hidden-state of the [CLS] token in the original and perturbed input (higher is better)}
\label{fig:pertubationIMDB_RoBERTa}
\end{figure*}

\begin{figure*}[h]
    \centering
            \includegraphics[width=0.9\textwidth]{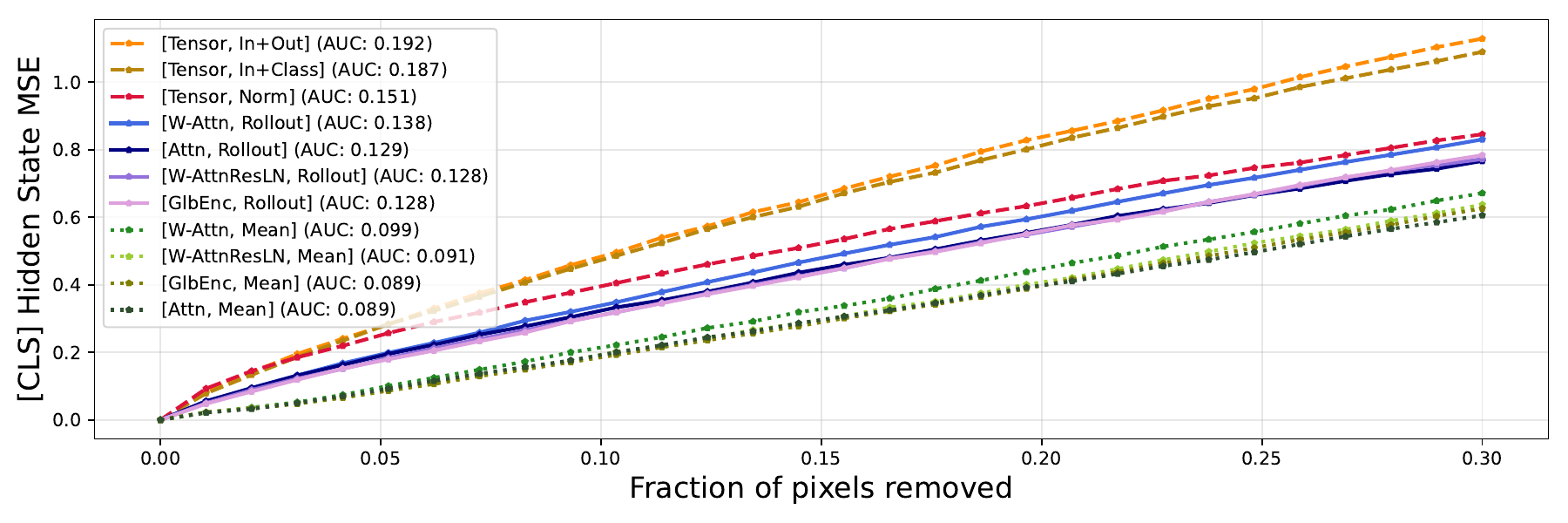}
    \caption{\textbf{Perturbation Tests in Vision:} Effect of perturbations on final hidden representations of DeiT-Small.}
    \label{fig:pertubationVision_DEIT_SMALL}
\end{figure*}

In addition to the perturbation tests presented in Section~\ref{subsec:pertubationTests}, this section presents further experiments with RoBERTa, DeiT-small, and the more recent ModernBert and Gemma3. The results are shown in Figures~\ref{fig:pertubationIMDB_RoBERTa} and~\ref{fig:pertubationVision_DEIT_SMALL}, respectively. As illustrated, across all benchmarks, our method achieves higher AUC scores, consistently outperforming the baselines for all perturbation fractions. Furthermore, in Table~\ref{tab:Pythia-1B}, we present perturbation results for decoder-only models, while Table~\ref{tab:bert_gemma} reports results for more modern models, including ModernBert~\cite{modernbert} and Gemma3~\cite{team2025gemma}.

%%%%%%%%%%%%%%%%%

%%%%%%%%%%%%%%%%%%%

% file: table_results.tex

\begin{table*}[h]
\centering
\small
\resizebox{\textwidth}{!}{
\begin{tabular}{ll|ccc|cccc|cccc}
\toprule
  &  Method: & \multicolumn{3}{c}{\textbf{Tensor (ours)}}& \multicolumn{4}{c}{\textbf{Rollout over layers}}& \multicolumn{4}{c}{\textbf{Mean over heads \& layers}}\\
\toprule
 LLM & Metric& Norm& In+Out& In+Class& Attn& W-Attn& W-AttnResLN& GlbEnc%1+2
& Attn& W-Attn& W-AttnResLN& GlbEnc %&Ordered
\\
\midrule
\textbf{Pyth}& HS-MSE $\uparrow$
& \textbf{3.708}& 3.603& \underline{3.683}& 0.306& 0.213& 0.233& N.A& 3.483& 3.650& \textbf{3.708}& N.A \\
                & AOPC $\uparrow$
                & \underline{0.147} & 0.146  
& \underline{0.147}  & 0.044  & 0.035  & 0.037  & N.A  & 0.141  & \underline{0.147}  & \textbf{0.148} & N.A  \\
\midrule
\textbf{Pico}& HS-MSE $\uparrow$

& 0.222&  \underline{0.223}& 0.22& 0.03& 0.014& 0.090& 0.036& 0.211& 0.22& 0.222& \textbf{0.225}\\
                 & AOPC $\uparrow$
                 & \textbf{0.129}& \underline{0.128}& \textbf{0.129}& 0.019& 0.020& 0.067& 0.036& 0.125& 0.127& \textbf{0.129}& \textbf{0.129}\\
\midrule
\textbf{Phi}& HS-MSE $\uparrow$
& 0.892& 0.887& 0.886& 0.079& 0.067& 0.053& N.A& 0.864& \underline{0.905}& \textbf{0.934}& N.A\\
              & AOPC $\uparrow$
              & 0.141& \underline{0.143}& \underline{0.143}& 0.028& 0.023& 0.026& N.A& 0.133& 0.141& \textbf{0.144}& N.A\\

\bottomrule
\end{tabular}
}
\caption{\textbf{Next Token Prediction Perturbation.} Results are AUC (higher is better) of (i) HS-MSE: Mean squared error between the last hidden-state of the final token in the original and perturbed input. (ii) AOPC: Absolute difference of the soft-maxed probability of the original predicted token, between the original and perturbed input. The GlbEnc results are not presented for the Pythia and Phi models, since their method is inapplicable for parallel-residual architectures. 'Pyth' for Pythia.}
\label{tab:Pythia-1B}
\end{table*}
%%%%%%%%%%%
%graphs layout:
%vision- deit  small\base hs graph in main
%       - deit small\base AOPC in appendix as graph\table

%imdb  - bert\roberta hs graph in main
%       - bert\roberta AOPC in appendix as graph\table

%wiki  - regular results in main as table.
%       - maybe result without two first tokens in appendix
% file: table_results.tex

\begin{table*}[t]
\centering
\small
\resizebox{\textwidth}{!}{
\begin{tabular}{ll|cc|cccc|cccc}
\toprule
  &  Method: & \multicolumn{2}{c}{\textbf{Tensor (ours)}}& \multicolumn{4}{c}{\textbf{Rollout over layers}}& \multicolumn{4}{c}{\textbf{Mean over heads \& layers}}\\
\toprule
 LLM & Metric& Norm& In+Out&  Attn& W-Attn& W-AttnResLN& GlbEnc%1+2
& Attn& W-Attn& W-AttnResLN& GlbEnc %&Ordered
\\
\midrule
\textbf{ModernBert}& HS-MSE $\uparrow$
& \underline{0.081} & \textbf{0.112} & 0.05 & 0.066 & 0.064 & 0.063 & 0.069 & 0.072 & 0.077 & 0.075\\

\midrule
\textbf{Gemma3}& HS-MSE $\uparrow$

& \underline{0.029} & \textbf{0.049} & 0.014 & 0.014 & 0.019 & 0.017 & 0.023 & 0.024 & 0.02 & 0.021\\

% \midrule
% \textbf{Phi}& HS-MSE %$\uparrow$
% & 0.892& 0.887& 0.886& 0.079& 0.067& 0.053& N.A& 0.864& \underline{0.905}& \textbf{0.934}& N.A\\
%               & AOPC %$\uparrow$
%               & 0.141& \underline{0.143}& \underline{0.143}& 0.028& 0.023& 0.026& N.A& 0.133& 0.141& \textbf{0.144}& N.A\\

\bottomrule
\end{tabular}
}
\caption{\textbf{Perturbation Tests in NLP with Modern Models:} Effect of token perturbations on final hidden representations of ModernBert-Base and Gemma3-270M, trained for sentiment prediction on the IMDB dataset. Results are AUC of HS-MSE: Mean squared error between the last hidden-state of the final token in the original and perturbed input (higher is better).}
\label{tab:bert_gemma}
\end{table*}
%%%%%%%%%%%
%graphs layout:
%vision- deit  small\base hs graph in main
%       - deit small\base AOPC in appendix as graph\table

%imdb  - bert\roberta hs graph in main
%       - bert\roberta AOPC in appendix as graph\table

%wiki  - regular results in main as table.
%       - maybe result without two first tokens in appendix

\section{Tensor Derivation with Biases}
\label{sec:app_biases}
Here we reiterate the tensor derivation introduced in Section~\ref{sec:methods} while including the transformer weight biases. The perturbation experiments in Section~\ref{subsec:pertubationTests} use the tensor without biases as described in Section~\ref{sec:methods}, and the relation decoding experiments in Section~\ref{subsec:arpproxLinear} use the full affine transformation described here.

We denote the model biases as $B\in\mathbb{R}^{L\times D}$, broadcasting the original $b\in\mathbb{R}^{D}$ biases to each sequence position $L$. For each module $f$ in the transformer block, we get an affine transformation of the form:
\begin{equation}\nonumber
\mathrm{vec}\!\bigl[f(X)\bigr]=\mathcal{T}^{(f)}\mathrm{vec}[X]+\mathrm{vec}[B^{(f)}]
\end{equation}
\paragraph{Tensorized Self-Attention.}
For an input $X\in\mathbb{R}^{L\times D}$, multi-head self attention is defined by:
\begin{equation}\nonumber
\mathrm{Attn}(X)=\sum_{h=1}^{H}A_{h}\,\left(X\,W_{v,h}+B_{v,h}\right)\,W_{o,h}+B_{o,h}
\end{equation}
\begin{equation}\nonumber
=\sum_{h=1}^{H}A_{h}\,\left(X\,W_{v,h}\right)\,W_{o,h}+B_{\text{attn}}\,,
\end{equation}
where $B_{\text{attn}}=B_{v,h}W_{o,h}+B_{o,h}$. The biases of the query and key projections are absorbed in the attention matrix $A_h$. 
Vectorising and grouping heads gives the attention tensor ${\mathcal{A}}$:
\begin{equation}\nonumber
\mathrm{vec}[\mathrm{Attn}(X)]=\underbrace{\sum_{h=1}^{H}\!\left(\left(W_{v,h}W_{o,h}\right)^{\top}\otimes A_{h}\right)}_{{\displaystyle \mathcal{A}}}\mathrm{vec}[X]
+\mathrm{vec}[B_{\text{attn}}]\in\mathbb{R}^{LD} \,.
\end{equation}
where $\mathcal{A}\in\mathbb{R}^{LD\times LD}$ is flattened to a matrix as defined in Eq.~\eqref{eq:preq_mat_tensor}.

\paragraph{Tensorized LayerNorm.} 
With weights $\gamma\in\mathbb{R}^{D\times D}$ and bias $\beta\in\mathbb{R}^{L\times D}$, the LayerNorm
\begin{equation}\nonumber
\mathrm{LayerNorm}(X)=\gamma\!\odot\!\tfrac{X-\mu}
{\sqrt{\sigma^{2}+\varepsilon}}+\beta \,,
\end{equation}
is similarly vectorized as:
\begin{equation}\nonumber
\mathrm{vec}\!\bigl[\mathrm{LayerNorm}(X)\bigr] = 
\underbrace{\bigl[(I_{D}-\tfrac{\mathbf{1}\mathbf{1}^{\!\top}}{D})
      \mathrm{ diag} (\gamma)\bigr]^{\!\top}
      \!\otimes\!\mathrm{ diag }\!\bigl(\tfrac{1}{\sqrt{\sigma^{2}+\varepsilon}}\bigr)}_{\displaystyle\mathcal{L}}
      \mathrm{vec}[X]
+\mathrm{vec}[\beta]\in\mathbb{R}^{LD} \,.
\end{equation}

\paragraph{Tensorized FFN}
Given an activation function $\phi$, the FFN is defined by two linear layers as follows:
\begin{equation}\nonumber
\mathrm{FFN}(X) = \phi(XM_{1}+B_{M_1})M_{2}+B_{M_2}.  
\end{equation}

The element-wise activation can be converted to an input-dependent hadamard product $\frac{\phi(Z)}{Z}\odot Z$, and tensorized as:
\begin{equation}\nonumber
\mathrm{vec}\!\bigl[\phi(Z)\bigr] =\underbrace{\mathrm{diag}\!\left(\mathrm{vec}\left[\frac{\phi(Z)}{Z}\right]\right)}_{{\displaystyle \Psi}}\mathrm{vec}[Z]\,.
\end{equation}

Resulting in the full vectorized form of the FFN as follows:
\begin{equation}\nonumber
\mathrm{vec}\!\bigl[\mathrm{FFN}(X)\bigr]=\underbrace{\bigl(M_{2}^{\top}\otimes I_{L}\bigr)\Psi\bigl(M_{1}^{\top}\otimes I_{L}\bigr)}_{{\displaystyle \mathcal{M}}}\mathrm{vec}[X]\,
+\underbrace{\mathrm{vec}[B_{M_2}]+\mathrm{\bigl(M_{2}^{\top}\otimes I_{L}\bigr)\Psi vec}[B_{M_1}]}_{\mathrm{vec}[B_{\text{FFN}}]}\in\mathbb{R}^{LD} \,.
\end{equation}
which is characterized by a tensor $\mathcal{M}$.

\paragraph{Transformer Block as Tensor.}
As defined in Eq.~\eqref{eq:BlockAtTensor}, stacking the tensors obtained from the self-attention, residual, normalizations, produces the tensor $\mathcal{T}^n$ associated with the $n$-th post-layernorm block:
 \begin{equation}\nonumber
\mathcal{T}^{n}=\mathcal{L}_{2}^{n}\left(\mathcal{M}^{n}+\mathcal{I}\right)\mathcal{L}_{1}^{n}\left(\mathcal{A}^{n}+\mathcal{I}\right)+\mathrm{vec}[B_{\text{block}}^{n}]
\end{equation}

Where the bias of each sub-module is transformed by the following ones as:
\begin{equation}\nonumber
\mathrm{vec}[B_{\text{block}}^{n}]=\mathrm{vec}[\beta_{2}^{n}]
+\mathcal{L}_{2}^{n}\left(\mathrm{vec}[B_{\text{FFN}}^{n}]+\mathcal{M}^{n}\left(\mathrm{vec}[\beta_{1}^{n}]+\mathcal{L}_{1}^{n}\mathrm{vec}[B_{\text{attn}}^{n}]\right)\right)
\end{equation}
The derivation for a pre-layernorm block is obtained similarly, by changing the order of the components:
 \begin{equation}\nonumber
\mathcal{T}^{n}=\left(\mathcal{I}+\mathcal{M}^{n}\mathcal{L}_{2}^{n}\right)\left(\mathcal{I}+\mathcal{A}^{n}\mathcal{L}_{1}^{n}\right)+\mathrm{vec}[B_{\text{block}}^{n}]
\end{equation}

\paragraph{Entire Transformer as Tensor.} 
As shown in Eq.~\eqref{eq:transformer_as_tensor}, the entire model $\mathcal{F}$ is fully represented as a chain of high-order tensor transformations. Adding biases results in the final affine transformation:
\begin{equation}\nonumber
\mathrm{vec}[\mathcal{F}(X)] = \left( {\prod_{n=1}^N \mathcal{T}^{n}} \right) \mathrm{vec}\bigl[X\bigr] + \mathrm{vec}[B_{\mathrm{full}}]\,,
\end{equation}
 where the bias of each block is recursively transformed by the following ones:
\begin{equation}\nonumber
\mathrm{vec}[B_{\mathrm{full}}]=\mathrm{vec}[B_{\text{block}}^{N}]+\mathcal{T}^{N}\left(\mathrm{vec}[B_{\text{block}}^{N-1}]+\cdots\right)\,.
\end{equation}
 
%%%%%%%%%%%%%%%%%%%%%%%%%%%%%%%%%%%%%%%%%%%%%

\section{Memory-Efficient Tensor Computation}
\label{sec:appendix_mem_eff_compute}
The tensor computation introduced in Section~\ref{sec:methods} relies on multiplications of large matrices in $\mathbb{R}^{LD\times LD}$, which may be prohibitive for larger models and longer input sequences. In practice, we use a memory-efficient computation method based on the following observations. \textbf{(i)} Given an input $X$, patching the original Transformer function $\mathcal{F}$ to use the precomputed attention matrices, FFN activations, and LayerNorm variance from the forward pass on $X$ yields a linear function $\widetilde{\mathcal{F}}_X$ whose Jacobian is exactly the desired tensor:
\begin{equation}\label{eq:jacobian}
    \mathcal{T}_X=\frac{\partial \widetilde{\mathcal{F}}_X(X)}{\partial X}\in \mathbb{R}^{L\times D \times L \times D}.
\end{equation}
\textbf{(ii)} The tensor $\mathcal{T}_X\in \mathbb{R}^{L\times D \times L \times D}$ captures the influence of each input token-channel pair on every output token-channel pair. Since in both the input attribution and relation decoding experiments we are only interested in the influence on a single output position $\ell_{\mathrm{out}}\in[L]$ (either the last token or the \texttt{[CLS]} token), it suffices to compute only the 3-dimensional tensor slice corresponding to that position, i.e., $\mathcal{T}_{X[\ell_{\mathrm{out}},:,:,:]}\in \mathbb{R}^{D \times L \times D}$.

Thus, in our experiments we compute only this 3-d slice using the Jacobian of the patched Transformer function, as in Eq.~(\ref{eq:jacobian}).

If needed, the full 4-d Jacobian and tensor can be computed in a memory-efficient manner using forward-mode differentiation, such that the full tensor is never materialized on the GPU. This is done by applying the patched Transformer function $\widetilde{\mathcal{F}}_X$ to unit (basis) matrices $E^{\ell,d}\in\mathbb{R}^{L\times D}$,
\begin{equation}\nonumber
\big(E^{\ell,d}\big)_{i,j}=
\begin{cases}
1,& i=\ell \land j=d\\
0,& \text{else,}
\end{cases}
\end{equation}
such that
\begin{equation}
    \forall \ell\in [L],\ \forall d\in [D]:
    \ \mathcal{T}_{X[:,:, \ell,d]}=\widetilde{\mathcal{F}}_X\!\left(E^{\ell,d}\right).
\end{equation}
This allows computing the entire tensor using $L\cdot D$ (possibly batched) forward passes, trading GPU memory for compute time.

\section{Relation Decoding Experiment}
\label{sec:appendix_reltion_relation_decoding}
We mostly adopt the experimental setup and relations dataset introduced in \citet{hernandez2024linearity}, using the relation categories of \textit{bias}, \textit{common sense}, and \textit{factual}. Although, in order to adapt to our tensor method we introduce several changes: \textbf{(i)} In order to perform the mean tensor approximation in Eq ~\eqref{eq:mean_T_relation}, we must filter the samples within each relation to those of the most common token length. \textbf{(ii)} Due to limited academic computational resources, we evaluate on Pythia-1B, which is a smaller model than used by \citet{hernandez2024linearity}. \textbf{(iii)} Since we use a smaller LM, we further filter the test samples only to those in which the correct object is within the top-20 tokens predicted by the model. 

For each relation type we use $m=3$ training examples to compute the mean tensor, and the LRE weights of \citet{hernandez2024linearity}, and report results averaged on 6 random seeds of train-test splits. 

For the mean tensor calculation in Eq.~\eqref{eq:mean_T_relation}, we use the full affine transformation with biases described in Appendix \ref{sec:app_biases} to obtain:
\begin{equation}\nonumber 
\widetilde{\mathcal{T}_r}=\frac{1}{m}\sum_{i}(\mathcal{T}_{X_{i}}+B_i)\,.
\end{equation}
This is necessary to obtain an accurate approximation of the original model.

Although the LRE method was developed for extracting linear relations from intermediate hidden-state, we compare it to ours using the input embeddings for a legitimate comparison. This method was shown as an ablation in their work. Additionally, the LRE baseline requires an additional hyper-parameter $\beta$ which scales the Jacobian matrices in their method. We use their default value for  Pythia's GPT-NeoX architecture of $\beta=2.5$, which gave the best results in a grid-search of other proposed values in their repository.

%%%%%%%%%%%%%%%%%%%%%%%%%%%%%%%%%%%%%
% \sub{Additional Results}\label{subsec:app_rel_more_results}
% \begin{figure*}[t]
%     % \centering
%         \includegraphics[width=0.99\textwidth]{plots/relation/pythia-1b-all-layers-per-relation.pdf}

%     \caption{ \textbf{Relation Decoding Using Intermediate Hidden States:} Accuracy per single relation, while computing Tensor and LRE \cite{hernandez2024linearity} model approximations using varying initial layers (12 layers for Pythia-1B) as in Eq. ~\eqref{eq:mean_T_relation_int_layer}. Relation names are color coded by category: {\color{darkgreen}Bias}, {\color{blue}Common Sense}, {\color{purple}Factual}. Mean results for each of category are  shown in the main article in Figure \ref{fig:relation_layers_mean}.}
%     \label{fig:relation_layers_examples}
% \end{figure*}
%%%%%%%%%%%%%%%%%%%%%%%%%%%%%%%%%%%%%v

\section{Tensor Approximation Error Bound}
\label{sec:app_approx_bound}
Proposition~\ref{proposition:senitivity} states that the \textbf{approximation error} of the tensor $\mathcal{T}_X$ computed on input $X$, when evaluating the transformer function $\mathcal{F}$ at $\left(X+\epsilon\right)$ is bounded by:
\begin{equation}
{\left\Vert \mathcal{T}_{X}\left(X+\epsilon\right)-\mathcal{F}\left(X+\epsilon\right)\right\Vert _{2}}\,\leq
\left\Vert \mathcal{T}_{X}\right\Vert _{2}\left\Vert \epsilon\right\Vert _{2}+\left\Vert \mathcal{F}\left(X+\epsilon\right)-\mathcal{F}\left(X\right)\right\Vert _{2}\,.
\end{equation}
Here we prove this claim and provide the explicit bound on the spectral norm of the tensor $\left\Vert \mathcal{T}_{X}\right\Vert _{2}$, when flattened as a matrix in $\mathbb{R}^{LD\times LD}$.

First, using the full derivation with biases in Appendix~\ref{sec:app_biases}, for any input embedding $X\in\mathbb{R}^{L\times D}$ and perturbation $\epsilon\in\mathbb{R}^{L\times D}$ we have:
% \begin{equation}\nonumber
% \left\Vert \mathcal{T}_{X}\left(X+\epsilon\right)-\mathcal{F}\left(X+\epsilon\right)\right\Vert _{2}
% \end{equation}
% \begin{equation}\nonumber
% =\left\Vert \mathcal{T}_{X}\text{vec}\left[X+\epsilon\right]+\text{vec}\left[B_{X}\right]-\text{vec}\left[\mathcal{F}\left(X+\epsilon\right)\right]\right\Vert _{2}
% \end{equation}
% \begin{equation}\nonumber
% =\left\Vert \mathcal{T}_{X}\text{vec}\left[\epsilon\right]-\text{vec}\left[\mathcal{F}\left(X\right)-\mathcal{F}\left(X+\epsilon\right)\right]\right\Vert _{2}
% \end{equation}
% \begin{equation}
% \leq\left\Vert \mathcal{T}_{X}\right\Vert _{2}\left\Vert \epsilon\right\Vert _{2}+\left\Vert \mathcal{F}\left(X+\epsilon\right)-\mathcal{F}\left(X\right)\right\Vert _{2}
% \end{equation}
\begin{align}
&\left\Vert \mathcal{T}_{X}\left(X+\epsilon\right)-\mathcal{F}\left(X+\epsilon\right)\right\Vert _{2} 
\nonumber\\
&= \left\Vert \mathcal{T}_{X} \, \text{vec}\left[X+\epsilon\right] 
+ \text{vec}\left[B_{X}\right] 
- \text{vec}\left[\mathcal{F}(X+\epsilon)\right] \right\Vert \nonumber \\
&= \left\Vert \mathcal{T}_{X} \, \text{vec}\left[\epsilon\right] 
- \text{vec}\left[\mathcal{F}(X) - \mathcal{F}(X+\epsilon)\right] \right\Vert_{2} \nonumber \\
&\leq \left\Vert \mathcal{T}_{X} \right\Vert_{2} \left\Vert \epsilon \right\Vert_{2} 
+ \left\Vert \mathcal{F}(X+\epsilon) - \mathcal{F}(X) \right\Vert_{2}
\stepcounter{equation}\tag{\theequation}\label{eq:proof_first_bound}
\end{align}

To bound $\left\Vert \mathcal{T}_{X}\right\Vert _{2}$, we bound the spectral norm of the tensor of each sub-module of the transformer block when flattened as a matrix (as defined in Section~\ref{sec:methods}), and then combine the bounds within the block and across layers.

\paragraph{Self Attention.}
Denoting the combined value-output projection per head $W_{v,h}W_{o,h}$ as $W_{vo}^{h}$ we get:
\begin{align*}
    \left\Vert \mathcal{A}\right\Vert _{2}
    &= \left\Vert \sum_{h\in H}\left(W_{vo}^{h}\right)^{\top}\otimes A^{h} \right\Vert _{2} \\
    % &\leq \sum_{h\in H} \left\Vert W_{vo}^{h} \right\Vert _{2} \left\Vert A^{h} \right\Vert _{2} \\
    &\leq \sum_{h\in H} \left\Vert W_{vo}^{h} \right\Vert _{2} \sqrt{ \left\Vert A_{x}^{h} \right\Vert _{1} \left\Vert A_{x}^{h} \right\Vert _{\infty} } \\
    &\leq \sqrt{L} \sum_{h} \left\Vert W_{vo}^{h} \right\Vert _{2} 
    \stepcounter{equation}\tag{\theequation} \label{eq:proof_attn}
\end{align*}
% \begin{equation}\label{eq:proof_attn}
% \left\Vert \mathcal{A}\right\Vert _{2}=\left\Vert \sum_{h\in H}\left(W_{vo}^{h}\right)^{\top}\otimes A^{h}\right\Vert _{2}
% \end{equation}
% \begin{equation}\nonumber
% \leq\sum_{h\in H}\left\Vert W_{vo}^{h}\right\Vert _{2}\left\Vert A^{h}\right\Vert _{2}
% \end{equation}
% \begin{equation}\nonumber
% \leq\sum_{h\in H}\left\Vert W_{vo}^{h}\right\Vert _{2}\sqrt{\left\Vert A_{x}^{h}\right\Vert _{1}\left\Vert A_{x}^{h}\right\Vert _{\infty}}
% \end{equation}
% \begin{equation}\nonumber
% \leq\sqrt{L}\sum_{h}\left\Vert W_{vo}^{h}\right\Vert _{2}
% \end{equation}
\paragraph{FFN.} The tensor of the FFN block for input $X$ is defined as:
\begin{equation}\nonumber
\mathcal{M}=\bigl(M_{2}^{\top}\otimes I_{L}\bigr)\mathrm{diag}\!\left(\mathrm{vec}\left[\frac{\phi(X)}{X}\right]\right)\bigl(M_{1}^{\top}\otimes I_{L}\bigr)\,,
\end{equation}
with the element-wise activation function $\phi$. Standard choices of $\phi$ such as GELU and SilU follow $\phi(x)\leq x$, so we have:  
\begin{equation}\nonumber
\left\Vert \mathrm{diag}\!\left(\mathrm{vec}\left[\frac{\phi(X)}{X}\right]\right)\right\Vert _{2}=\left\Vert \frac{\phi(X)}{X}\right\Vert _{\infty}\leq1
\end{equation}
Overall for the whole FFN:
\begin{equation}\label{eq:proof_FFN}
\left\Vert \mathcal{M}\right\Vert _{2}\leq\left\Vert M_{2}\right\Vert _{2}\left\Vert M_{1}\right\Vert _{2}
\end{equation}
\paragraph{LayerNorm.} The LayerNorm tensor is defined as:
\begin{equation}\nonumber
\mathcal{L}_X=\bigl[(I_{D}-\tfrac{\mathbf{1}\mathbf{1}^{\!\top}}{D})\mathrm{diag}(\gamma)\bigr]^{\!\top}\!\otimes\!\mathrm{diag}\!\bigl(\tfrac{1}{\sigma_X}\bigr)
      \,,
\end{equation}

For the left side of the mean centering matrix and $\gamma$ we have:
\begin{equation}\nonumber
\left\Vert I_{D}-\tfrac{\mathbf{1}\mathbf{1}^{\!\top}}{D}\right\Vert _{2}\leq1\,,\,\left\Vert \mathrm{diag}(\gamma)\right\Vert _{2}=\left\Vert \gamma\right\Vert _{\infty}
\end{equation}
And for the right side of the variance $\sigma_X\in\mathbb{R}^{L\times L}$ we have:
\begin{equation}\nonumber
\left\Vert \mathrm{diag}\!\bigl(\tfrac{1}{\sigma_{X}}\bigr)\right\Vert _{2}=\left\Vert \tfrac{1}{\sigma_{X}}\right\Vert _{\infty}=\max_{l\in L}\frac{1}{\text{Var}\left[X_{\left[l,:\right]}\right]}
\end{equation}
Where $\text{Var}\left[X_{\left[l,:\right]}\right]$ is the variance of $X$ at position $l\in L$.
Importantly, this is the only data-dependent quantity in our bound, depending on the minimal variance of each of the hidden-states $X_{\left[l,:\right]}\in\mathbb{R}^D$ at the input to the layer norm.
We denote it as  
\begin{equation}\nonumber
\xi_{\left(\text{LN},X\right)}=\min_{l\in L}\text{Var}\left[X_{\left[l,:\right]}\right],
\end{equation}
and the overall bound for the LayerNorm tensor is:
\begin{equation}\label{eq:proof_LN}
\left\Vert \mathcal{L}_{x}\right\Vert _{2}\leq\frac{\left\Vert \gamma\right\Vert _{\infty}}{\xi_{\left(\text{LN},X\right)}}\, .
\end{equation}
\paragraph{Whole Transformer.} The tensor of a post-layernorm tramformer layer $n\in N$ is
 \begin{equation}\nonumber
\mathcal{T}^{n}=\mathcal{L}_{2}^{n}\left(\mathcal{M}^{n}+\mathcal{I}\right)\mathcal{L}_{1}^{n}\left(\mathcal{A}^{n}+\mathcal{I}\right).
\end{equation}
Combining the bounds of each component from Eq.~\eqref{eq:proof_attn},\eqref{eq:proof_FFN},\eqref{eq:proof_LN} we get: for the whole transformer:

\begin{align}
\left\Vert \mathcal{T}_{X} \right\Vert_{2}
&\leq \prod_{n=1}^{N}
\left\Vert \mathcal{L}_{2}^{n} \right\Vert 
\left( \left\Vert \mathcal{M}^{n} \right\Vert + 1 \right)
\left\Vert \mathcal{L}_{1}^{n} \right\Vert 
\left( \left\Vert \mathcal{A}^{n} \right\Vert + 1 \right) \nonumber \\
&\leq \prod_{n=1}^{N}
\frac{\left\Vert \gamma_{2}^{n}\right\Vert _{\infty}}{\xi_{(\text{LN}_{2}^{n},X)}}\left(\left\Vert M_{1}^{n}\right\Vert _{2}\left\Vert M_{2}^{n}\right\Vert _{2}+1\right) \nonumber \\
&\quad \cdot
\frac{\left\Vert \gamma_{1}^{n}\right\Vert _{\infty}}{\xi_{(\text{LN}_{1}^{n},X)}}\bigl(\sqrt{L}\sum_{h}\left\Vert W_{vo}^{h}\right\Vert _{2}+1\bigr)
\stepcounter{equation}\label{eq:proof_whole_model}
\end{align}

Together with Eq.~\ref{eq:proof_first_bound}, this completes the proof of Proposition~\ref{proposition:senitivity}.

We note that although this bound is data-dependent, the value of 
\begin{equation}\nonumber
\left\Vert \mathcal{L}_{x}\right\Vert _{2}\leq\frac{\left\Vert \gamma\right\Vert _{\infty}}{\xi_{\left(\text{LN},X\right)}}
\end{equation} 
is typically a small constant.

%%%%%%%%%%%%%%%%%%%%%%%%%%%%%%%%%%%%%%%%%%%%%%%%%%%%%%%%%%%%%%
\end{document}